\documentclass[12pt, conference]{IEEEtran}

\ifCLASSINFOpdf
   \usepackage[pdftex]{graphicx}
\fi
\usepackage[utf8]{inputenc}
\usepackage[cmex10]{amsmath}
\usepackage{booktabs}
\usepackage[dvipsnames]{xcolor}
\usepackage{xcolor}
\usepackage{array}

\DeclareUnicodeCharacter{00AD}{~}

\begin{document}

\title{Lightweight Monocular Depth Estimation}

\author{
\IEEEauthorblockN{Ruilin Ma}
\IEEEauthorblockA{\textit{Department of Computing Science, Multimedia} \\
\textit{University of Alberta}\\
Edmonton, Alberta, Canada T6G 2H1\\
ruilin3@ualberta.ca}
\and
\IEEEauthorblockN{Shiyao Chen}
\IEEEauthorblockA{\textit{Department of Computing Science, Multimedia} \\
\textit{University of Alberta}\\
Edmonton, Alberta, Canada T6G 2H1\\
shiyao1@ualberta.ca}

\and
\IEEEauthorblockN{\centerline{Qin Zhang}}
\IEEEauthorblockA{\textit{Department of Computing Science, Multimedia} \\
\textit{University of Alberta}\\
Edmonton, Alberta, Canada T6G 2H1\\
qin7@ualberta.ca}
}

\maketitle

\begin{abstract}
    Monocular depth estimation can play an important role in addressing the issue of deriving scene geometry from 2D images.
   It has been used in a variety of industries, including robots, self-driving cars, scene comprehension, 3D reconstructions, and others.
   The goal of our method is to create a lightweight machine-learning model in order to predict the depth value of each pixel given only a single RGB image as input with the Unet structure of the image segmentation network. We use the NYU Depth V2 dataset to test the structure and compare the result with other methods. The proposed method achieves relatively high accuracy and low root-mean-square error.
\end{abstract}  
\section{Introduction}

The process of estimating the distance of objects in a scene from a single image captured by a camera is known as monocular depth estimation. This is a critical problem in computer vision because it enables algorithms to understand the spatial layout of a scene, which is required for a variety of applications including robotics, augmented reality, and autonomous vehicles. One of the major challenges in monocular depth estimation is that a single image does not contain enough information to determine the depth of objects in a scene accurately. To overcome this challenge, algorithms must use other cues such as the relative size of objects, the relative position of objects, and the amount of occlusion between objects to estimate their distances from the camera. Hence, we convert the image depth estimation problems into image classification problems with UNet architecture and transfer learning. We will further explore the network in the Methods section.
\section{Literature review}
\subsection{Background Subtraction}
Background subtraction has been a popular yet challenging study area in computer vision recently. Background subtraction is a technique to separate the foreground from the background scene in the video\cite{wu2019background}. There are two main ways to tackle this problem, conventional methods, such as statistical models\cite{BOUWMANS201431}, cluster models and estimation models\cite{BOUWMANS201431}, and machine learning approaches. Therefore, this section introduces different solutions to the background subtraction problem, including statistical approaches and machine learning approaches.

\subsubsection{Background Modeling by Stability of Adaptive Features in Complex Scenes}
The multi-feature-based background model has a pivotal role nowadays in background-subtraction research. This paper has proposed a novel multi-feature-based background model, named the stability of adaptive feature (SoAF) model, that uses the stabilities of several characteristics in a pixel to adaptively assess their contributions to foreground detection\cite{yang2018background}. It introduced the method by stating the existing main challenges to background subtraction for dynamic and complex scenes. In addition, with the traditional method's comparison, the SoAF has the advantage of self-adaptability, complementary and robustness.
\subsubsection{Background Subtraction based on Deep Pixel Distribution Learning}
As traditional background subtraction negatively impacts scene variability, this paper introduced Deep Pixel Distribution Learning (DPDL), which automatically adjusts to the many scene-specific distributions\cite{zhao2018background}. With the methods of the Random Permutation of Temporal Pixels, convolutional neural network, and pixel-wise representation. In particular, the random permutation of temporal pixels is applied for pixel-wise representation. The proposed method concentrates on a more basic idea behind background subtraction, which is the categorization of pixels in a time sequence using network-learned discriminative features that directly affect the distribution of pixels across time.
\subsubsection{Background Subtraction Based on Integration of
Alternative Cues in Freely Moving Camera}
One method to subtract background is to integrate the alternative cues about foreground and background\cite{zhao2019background}. This approach uses the Gaussian Mixture model to estimate the foreground motion and uses spatiotemporal features filtered by homography transformation to capture the background cues. 

\subsubsection{Background Subtraction Based on Principal Motion for a Freely Moving Camera}
This paper introduces a novel method that solves the background subtraction problem for a freely moving camera. The angle and magnitude represent the optical flow, which is the motion of pixels. To distinguish between the motion of moving objects and background motion, Robust Principal Components Analysis (RPCA) is used\cite{ma2020background}. Besides, super-pixels are used to refine the foreground mask created by the revised RPCA. Even though the suggested method did not provide a flawless solution, it did offer a promising technique for approaching the issue's resolution.

\subsubsection{Background Subtraction by Difference Clustering}
Unlike K-means clustering algorithm, the novel difference clustering needs only two clusters, the background and foreground, using the quartile method, and one iteration to get the background subtracted from the scene\cite{wu2019background}. Because difference clustering requires only one iteration, it dramatically reduces the computational time that enables a real-time subtraction for background and foreground for given videos. The experiment result shows that the difference clustering has a higher Precision and F-measure value.  Moreover, the experiments demonstrate a better performance compared to other unsupervised background subtraction methods.

\subsubsection{Dynamic Deep Pixel Distribution Learning
for Background Subtraction}

The Dynamic Deep Pixel Distribution Learning (D-DPDL) model is a new approach for background subtraction. This model automatically learns the statistical distribution generated by Random Permutation of Temporal Pixels (RPoTP)\cite{zhao2020dynamic}. However, some noises might be produced by the random permutation. Therefore, this method takes the Bayesian refinement model as the solution to reduce the random noise. The novel part of this approach is that the D-DPDL model can perform well estimation even if the training video and testing video are entirely different since the distribution is a general feature no matter what the scene is.

\subsubsection{Deep Variation Transformation Network for Foreground Detection} 
The detection between foreground and background has a pivotal role in computer vision. However, traditional methods such as the Gaussian Mixture Model (GMM) and Kernel Density Estimation (KDE) tends to make wrong classification since there are several similarities between foreground and background regarding the observation of pixels. Therefore, the paper proposes a new foreground detection method called Deep Variation Transformation Network (DVTN), which targets on the implicit pattern for pixel variation\cite{ge2021deep}. The output of DVTN is a new transformation of pixel variations and in the final part, a linear classifier will label the pixel either foreground or background. The test suggests that the DVTN shows a better performance compared with other deep learning methods or traditional methods in complex natural scenes.

\subsubsection{Universal Background Subtraction Based on Arithmetic Distribution Neural Network} 
Another paper proposed a universal background subtraction method based on the Arithmetic Distribution Neural Network (ADNN), which learns the distributions of temporal pixels\cite{zhao2022universal}. A Bayesian refinement model focusing on neighboring information is implemented and a histogram of probability is also utilized. The result shows how effective this new framework based on arithmetic distribution operations is when compared to other techniques.

\subsubsection{Fused Geometry Augmented Images for Analyzing Textured Mesh} 
The paper proposed a texture and geometry-based multi-modal mesh surface representation\cite{Taha9191099}. The method was established to extend the application of deep-learning solutions to 3D data given in the form of triangular meshes, which were obtained by mapping between the 3D mesh domain and the 2D image domain. In contrast to existing methods, the purposed method does not rely on expensive tensor-based or multi-view inputs. As a result, this approach brings a positive impact on both memory and computation efficiency. 

\subsubsection{Realtime Background Subtraction from Dynamic Scenes}
This paper proposed an approach for real-time moving object detection, which focuses on dealing with the situation of change of background scene textures in video. It used concepts from the large margin principle and online learning to produce a generalization of the one-class support vector machines (1-SVMs) formalism. This method establishes the track of temporal changes and spatial relationships between adjacent pixels and enhances the efficiency with the use of highly parallel graphics processors (GPUs)\cite{cheng2009realtime}. The experiment shows the superiority of this novel real-time approach (over 80 frames per second) compared to other offline algorithms.

\subsection{U-Net}
U-Net is a powerful fully convolutional neural network for image segmentation. This network requires fewer training images but results in more precise segmentation of given images\cite{olaf2015unet}. This section introduces the basic architecture of the networks and their applications.

\subsubsection{U-Net: Convolutional Networks for Biomedical
Image Segmentation}

\begin{figure}
     \centering
     \includegraphics[width=0.4\textwidth,keepaspectratio]{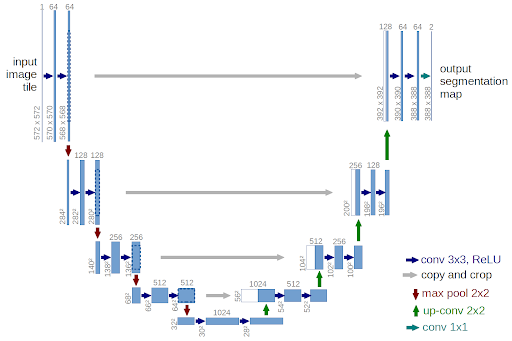}
     \caption{U-net architecture 32x32 pixels in the lowest resolution\cite{olaf2015unet}.}
     \label{unet}
\end{figure}

U-Net was initially designed for biomedical image segmentation in this paper. U-Net uses limited annotated samples but efficiently produces better segmentation results; see Figure \ref{unet}. The architecture shows a symmetric U-shape that consists of a contracting path and an expansive path\cite{olaf2015unet}. Since the training data are limited, excessive data augmentation that applies elastic deformation is used. With this data augmentation method, the deep learning model is able to learn invariance to such deformations. Because of this characteristic of the model, the U-Net can be applied to some medical fields for biomedical segmentation.

\subsubsection{3D U-Net: Learning Dense Volumetric Segmentation from Sparse Annotation} This paper proposed a network that extended the traditional U-Net by replacing the 2D operations with the 3D counterparts to do the dense volumetric segmentation with limited annotated 2D slices for training
\cite{cicek20163d}. The basic architecture of the 3D U-Net consists of a contracting encoder and an expanding decoder as the 2D U-Net. However, the innovational 3D U-Net takes in 3D volumes as input and processes them with 3D convolutions. This new network shows great potential in many biomedical applications since biomedical data are mostly in a 3D format, and it is very difficult and time-consuming to annotate a 3D object slice by slice. Therefore, with 3D U-Net, a 3D segmentation model will be created with very few training images.

\subsubsection{UNet++: A Nested U-Net Architecture for Medical Image Segmentation}
\begin{figure}
     \centering
     \includegraphics[width=0.4\textwidth,keepaspectratio]{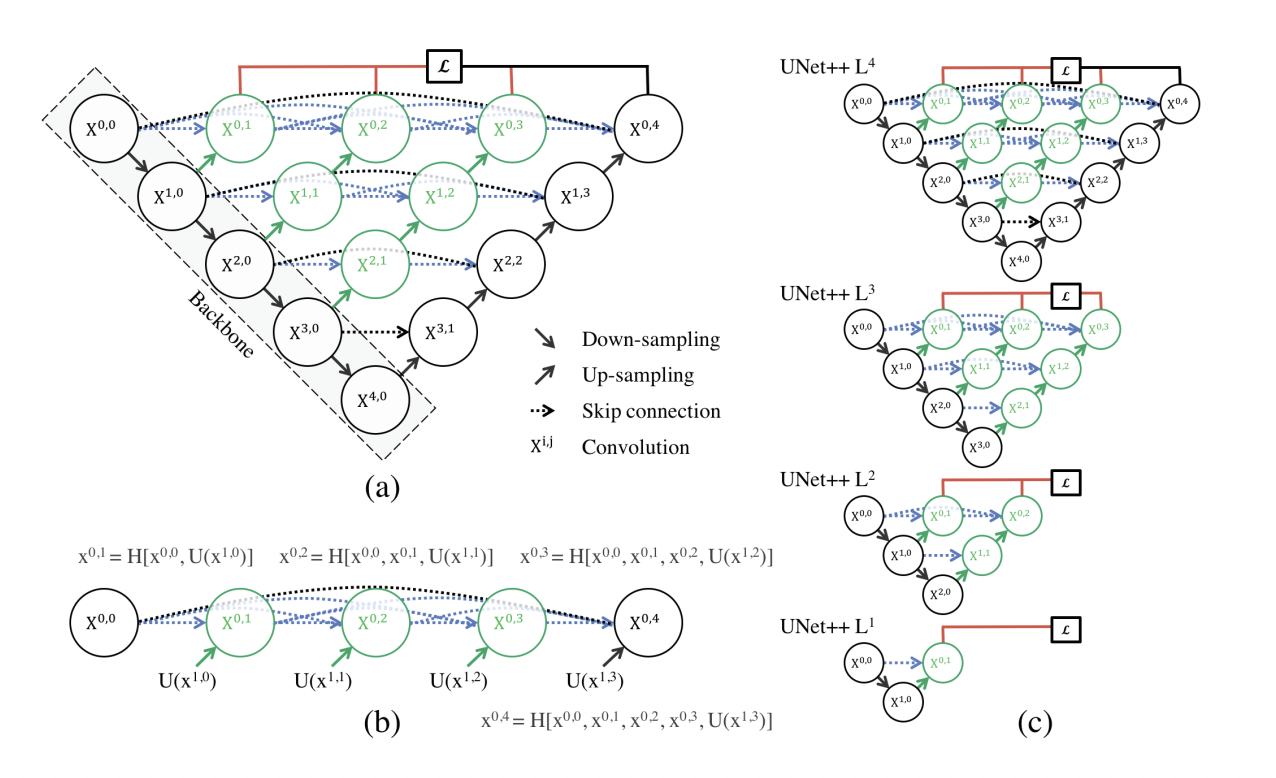}
     \caption{(a) UNet++ consists of an encoder and decoder that are connected through a series of nested dense convolutional blocks. (b) Detailed analysis of the first skip pathway of UNet++. (c) UNet++ can be pruned at inference time if trained with deep supervision\cite{zhou2018unet}.}
     \label{unetpp}
\end{figure}

This paper proposed a new deep learning network, UNet++, which is an improvement of the original U-Net. The novel architecture connects the encoder and decoder through a series of nested and dense skip pathways as Figure \ref{unetpp} shows. The nested pathways reduced the semantic gap between the feature maps of the encoder and decoder sub-networks. In consequence, the model can capture the foreground object at a better level. The improved architecture has shown better performance on nodule segmentation in the low-dose CT scans of the chest, nuclei segmentation in the microscopy images, liver segmentation in abdominal CT scans, and polyp segmentation in colonoscopy videos\cite{zhou2018unet}.

\subsubsection{Domain Adaptive Fusion for Adaptive Image Classification}

The research on deep adaptation networks has brought significant improvements on the domain domain classification tasks. The domain adaptive fusion has been proposed in the paper, a unique domain adaptation technique that, while being trained with a domain adversarial signal, promotes a domain-invariant linear connection between the pixel spaces of various domains and the prediction spaces\cite{dudley2020domain}. Instead of implementing traditional supervised learning methods, the performance of such a model may suffer rapid degradation when the target alters with training. They involved unsupervised domain adaption approaches for better performance.

\subsection{Transformer}
The attention-based model was initially proposed in computer vision field. Google Mind later made the Attention mechanism popular by publishing a paper that used an RNN model and added the Attention mechanism for image classification\cite{volodymyr2014recurrent}. After that the attention mechanism was then applied to the natural language processing. Bahdanau et al. introduces a sequence to sequence encoder-decoder architecture that shows a improvement on machine translation\cite{bahdanau2014neural}. Hence, this section will introduce the transformer architecture and its attention-based model.

\subsubsection{Attention Is All You Need}
This paper first introduces the transformer architecture based only on attention mechanisms, which completely abandoned network structure such as RNN and CNN.

\begin{figure}
     \centering
     \includegraphics[width=0.4\textwidth,keepaspectratio]{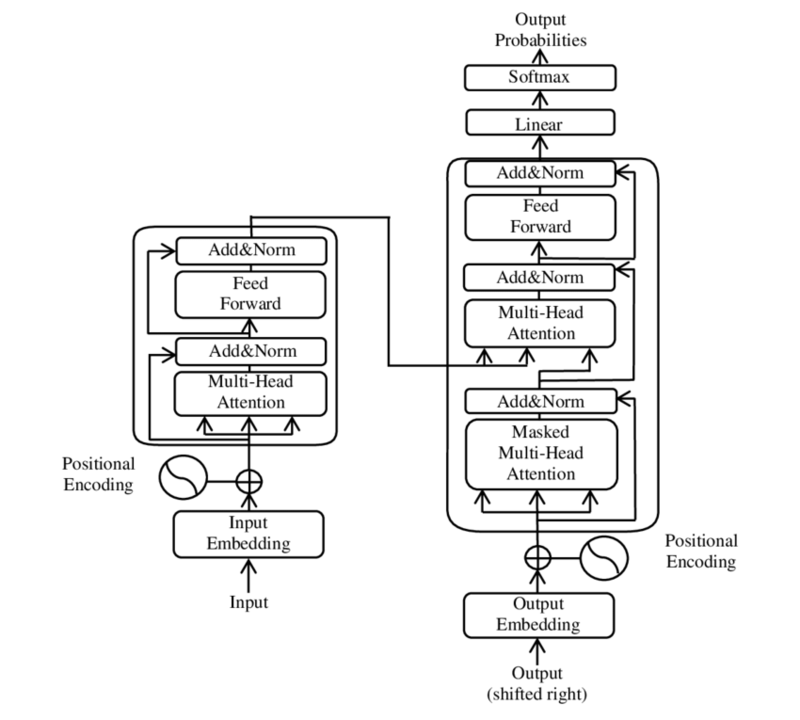}
     \caption{Transformer model architecture.}
     \label{transformer}
\end{figure}

Transformer follows the architecture of the encoder-decoder structure. However, it uses stacked self-attention and point-wise, fully connected layers for both the encoder and decoder\cite{ashish2017attention}, shown in the Figure \ref{transformer}. This novel architecture demonstrates excellent quality on machine translation tasks requiring significantly less time to train.

\subsubsection{An Image is Worth 16x16 Words: Transformers for Image Recognition at Scale}
This paper applies the standard Transformer structure, which was usually used in natural language processing, to image classification. The image was split into several patches, and put the linear sequence of those patches into the transformer. Therefore, the images can be trained as tokens in NLP\cite{alexey2020an}. This application shows a great result when the training dataset are large, but poorer performance compared to some powerful convolutional neural networks when the training sets are relatively small.

\subsubsection{Is Space-Time Attention All You Need for Video Understanding?}
The paper focuses on the research of migrating self-attention to the video domain, and proposes a self-attention model TimeSformer. The model separatively uses self-attention to model video data from both spatio-temporal latitudes\cite{gedas2021is}. Eventually, this model achieves the best performance on video classification among all the models, and demonstrates a better capability to learn from longer video clips compared to traditional 3D convolutional networks.

\subsubsection{Swin-Depth: Using Transformers and Multi-Scale Fusion for Monocular-Based Depth Estimation} 
An essential and fundamental job in computer vision is depth estimation using monocular sensors. Due to the limitation of convolutional neural networks' inherent operation, this paper has purposed a Swin-transformer method, a monocular depth estimation technique based on a transformer that uses hierarchical representation learning with linear complexity for pictures. The Swin-transformer will function as a backbone network and enhance the efficiency of monocular depth estimation with the help of its superior local modeling features. In addition, after learning about hierarchical representations, a simple attention module based on multi-scale fusion can help improve global knowledge learning. The unnecessary parameters in the monocular depth estimate utilizing the transformer are successfully reduced by the purpose technique\cite{cheng2021swin}.

\subsubsection{Pooling Pyramid Vision Transformer for Unsupervised Monocular Depth Estimation}
The majority of efforts made in monocular depth estimation are still based on convolutional neural networks, which have defects in high-resolution features and full-stage global information when extracting multi-scale features. The paper has purposed a new method called pooling pyramid vision transformer (PPViT), which method significantly narrows down the multi-scale features and reduces the sequence length for attention operations\cite{zhang2022pooling}. Compared with previous transformer-based work, the PPViT method has exhibited decreased latency.

\subsection{Depth Estimation}
Research on depth estimation has a long tradition in computer vision. A dense depth of map that estimates the absolute, even the relative depth in the real world is fundamental to lots of applications such as navigation and image segmentation\cite{clement2018digging}. In this section, we introduce some solutions for depth estimation.

\subsubsection{Self-organizing background subtraction using color and depth data} 
The RGBD-SOBS algorithm is presented in this paper to detect moving objects in RGBD videos. A self-organizing neural background model is also adopted to separately model the color and depth background. This paper used the resulting color and depth detection masks to guide the update of the model and the final result demonstrates that utilizing depth information rather than just color allows for significantly better performance. Besides, the comparison result with other existing methods also shows that  RGBD-SOBS has the highest average rating\cite{Maddalena2018SelforganizingBS}.

\subsubsection{Deep Learning based Monocular Depth Prediction: Datasets, Methods and Applications}

The paper has given a review of deep learning-based monocular depth prediction from three perspectives: supervised learning-based methods, unsupervised learning-based methods, and sparse sample guidance-based methods. Among other depth estimation methods, this paper emphasizes the role of deep learning and machine learning in monocular estimation. With a review of each method's advantage\cite{qing2020deep}, the authors also stated the challenge as well as the future trends of these methods.

\subsubsection{Mixed-Scale Unet Based on Dense Atrous Pyramid for Monocular Depth Estimation}

In the realm of image processing, monocular depth estimation is growing in popularity. In this study, the authors proposed a coder-decoder-based dense atrous pyramid-based mixed-scale Unet network (MAPUnet). Additionally, they debuted the Unet ++ segmentation network. As a replacement in the middle transducer part with a dense lustrous pyramid structure has been made, the purposed deep neural network-based framework model can avoid manual feature extraction. In addition, this network has fewer weights than the prior monocular estimation network\cite{yang2021mixed}.

\subsubsection{Depth extended online RPCA with spatiotemporal constraints for robust background subtraction}
To achieve strong background subtraction, this paper extends Online Robust Principal Component Analysis (OR-PCA) by incorporating depth and color information. Shadows and the similarity of color between the background and foreground objects have less impact on depth. When the foreground object is far from the camera's vision, it might not be captured because depth performs worse without color information. The integration of spatiotemporal constraints also contributes to the robustness of the new model\cite{javed2015depth}. The suggested strategy outperforms existing methods in experimental assessments with the use of color and range information. 

\subsubsection{Computationally efficient background subtraction in the light field domain}

In this paper, a brand-new method for background removal and depth estimation in light field photos is proposed. This approach revised Radon Transform without massive matrix calculations. Therefore, it is exceedingly computationally efficient and suitable for real-time use. To obtain a basic depth map of the captured image, this method takes advantage of the intrinsic structure and uniformity of the light field signal. The precise isolation of the background in the image is achieved by adopting a final segmentation method, which uses the extracted depth map as input\cite{ghasemi2014computationally}.   

\subsubsection{Look Deeper into Depth: Monocular Depth Estimation with Semantic Booster and Attention-Driven Loss}
Monocular depth estimations are greatly benefited from learning-based systems. By examining the training data, the research discovers that the per-pixel depth values in the present datasets frequently exhibit a long-tailed distribution. The performance of the model is constrained, especially in remote places, because most prior methods treated all areas in the training data equally despite the unequal depth distribution. In order to more effectively use the semantic information for the monocular depth estimation, scientists also designed a synergy network to automatically learn the information-sharing mechanisms between the two tasks\cite{jiao2018look}.

\subsubsection{TransformerFusion: Monocular RGB Scene Reconstruction using Transformers}
A novel transformer-based approach called TransformerFusion is proposed in this paper. A transformer network processes the frames from a monocular RGB video that turns the observation into feature fusion. The transformer architecture is essential to our methodology because it allows the network to learn to focus on the most informative view features for each 3D location in the scene\cite{bozic2021transformerfusion}. In comparison to other techniques, such as fully-convolutional 3D reconstruction, this method reduces runtime as well as enhances the performance of surface reconstruction.

\subsubsection{Dynamic Guided Network for Monocular Depth Estimation}
As a combination of self-attention mechanism and encoder-decoder, the author has purposed the DGNet method which enhances EMANet by including a powerful decoder module to gradually modify the coarse depth map. The purpose encoder may capture long-range information as well as assist with an efficient decoding module that employs a dynamic guided filter. The authors develop dynamic guided upsampling using spatially detailed data from low-level features to gradually improve depth maps. The suggested method provides depth maps with the appropriate level of precision, crisp structural features, and gradual depth shifts.\cite{xing9413264}.
\section{Method}

We propose the Unet network with a DenseNet-121 encoder and the basic blocks of convolutional layers as the decoder. The convolutional neural network computes a high-resolution depth map given a single RGB image with the help of transfer learning. 
\subsection{Network Architecture}
\begin{figure*}
  \includegraphics[width=\textwidth,height=4cm]{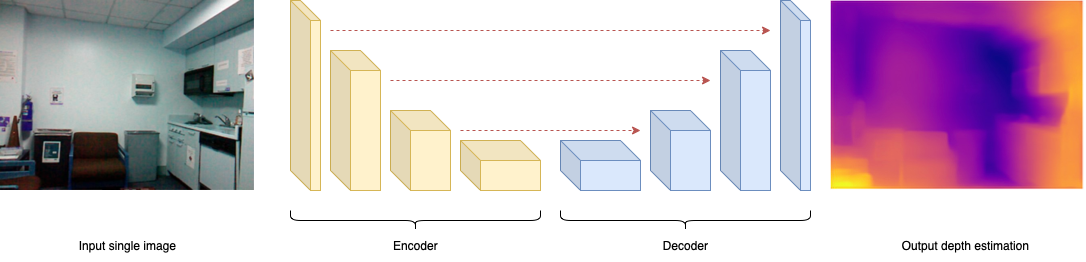}
  \caption{Overview of our U-net architecture.}
  \label{unet_arch}
\end{figure*}

Fig. \ref{unet_arch} illustrates the architecture of our encoder-decoder network. Because of the significant performance of DenseNet-121\cite{DBLP:journals/corr/HuangLW16a} on image classification, a pretrained DenseNet-121 model is used as the backbone of the U-Net, and it takes a single image as the input of the network. While the decoder uses basic blocks of up-sampling layers and associated skip-connections.

\subsection{Loss Function}
The loss function calculates the difference between the ground truth image depth $y$ and the predicted image depth $\hat{y}$. The loss function focuses on reducing the distortions of the image and minimizing the error of the depth values. The loss function $L$ consists of three loss functions:
\[
    L(y, \hat{y}) = \lambda L1 + L1_{grad} + L_{SSIM}
\]

L1 loss function, which is known as Least Absolute Deviations, calculates the sum off all the absolute differences between the ground truth values and the predicted values.

\[
    L1 = \frac{1}{n}\sum_{i=1}^n |y_i - \hat{y_i}|
\]

The second term of the loss function is the L1 loss defined over
the image gradient $g(y,\hat{y})$ of the depth image for both $x$ and $y$ coordinates:

\[
    L1_{grad} = \frac{1}{n}\sum_{i=1}^n (|g_x(y_i,\hat{y_i})| + |g_y(y_i,\hat{y_i})|)
\]
, where $g_x$ and $g_y$ are the L1 loss for both $x$ and $y$ directions.

Lastly, $L_{SSIM}$ uses the Structural Similarity (SSIM) \cite{zhou2004}. SSIM is a commonly-used metric for measuring the similarity between two images and is usually used on image reconstruction problems. The SSIM loss $L_{SSIM}$ is defined as:

\[
    L_{SSIM} = \frac{1 - SSIM}{2}
\]
, where SSIM defines as follow:
\[
    SSIM(x,y) = \frac{(2\mu_x\mu_y + c_1)(2\sigma_{xy} + c_2)}{(\mu^2_x +\mu^2_y+c_1)(\sigma^2_x +\sigma^2_y + c_2)}
\]
The weight parameter $\lambda$ only applies to the $L1$ loss function and we set it equals to 0.1.

When ground truth depth is bigger, the loss function will be also bigger. Therefore, we can apply the reciprocal of the depth to the original depth $y$ = $m / y$ \cite{Benjamin_2016}\cite{huang2018}, where m is the maximum depth of the image and $m = 10$ meters for the NYU V2 dataset.

\subsection{Dataset}

The NYU-Depth V2 dataset \cite{Silberman:ECCV12} consists of video sequences from various indoor scenes captured by Microsoft Kinect's RGB and Depth cameras. With 1449 densely labeled pairs of aligned RGB and depth images as well as 464 new scenes taken from 3 cities, images of this dataset have a resolution of 640 x 480 resolution. To close the gap in depth images, the depth photos are in-painted and positioned in relation to the RGB image. Among all 464 scenes, 249 scenes are used for training and 215 scenes for testing. To simplify the calculation process, we applied a reduced NYU-V2 dataset, and the simplified dataset was taken from \cite{alhashim_2022_ialhashimdensedepth}. The reduced version of the dataset contains 284 scenes with total of 50688 images for training. 
\subsection{Implementation Details}
The program was deployed with TensorFlow 2.6.2 with CUDA version 11.2 environment. The model was trained on 10 Epochs with a single RTX 2080 GPU for about 16 hours.

\subsection{Features}

Our project's important component is the lightweight UNet architecture, with DenseNet-121 \cite{DBLP:journals/corr/HuangLW16a} serving as the encoder. We select DenseNet-121 over other DenseNet models because of its relative lightness (Comparison details shown in Fig. 5.\cite{amaarora_2020}). There are only 8 million total parameters. At the same time, our lightweight model can still preserve high accuracy for the output result. Compared with the DenseNet-169 model with 14 Million total parameters, the advantage of DenseNet-121 is more obvious. This study offers fresh perspectives on how to increase the training model's effectiveness while using less training time. The preprocessing of input photos makes the most contribution to speeding up the model. Preprocessing is divided into two sections. The first step is converting the intensity value of each pixel point between (0,1). A further component flips the current image with a probability of 0.5 through image augmentation for generalization. 

\begin{figure*}
     \centering
     \includegraphics[width=13cm]{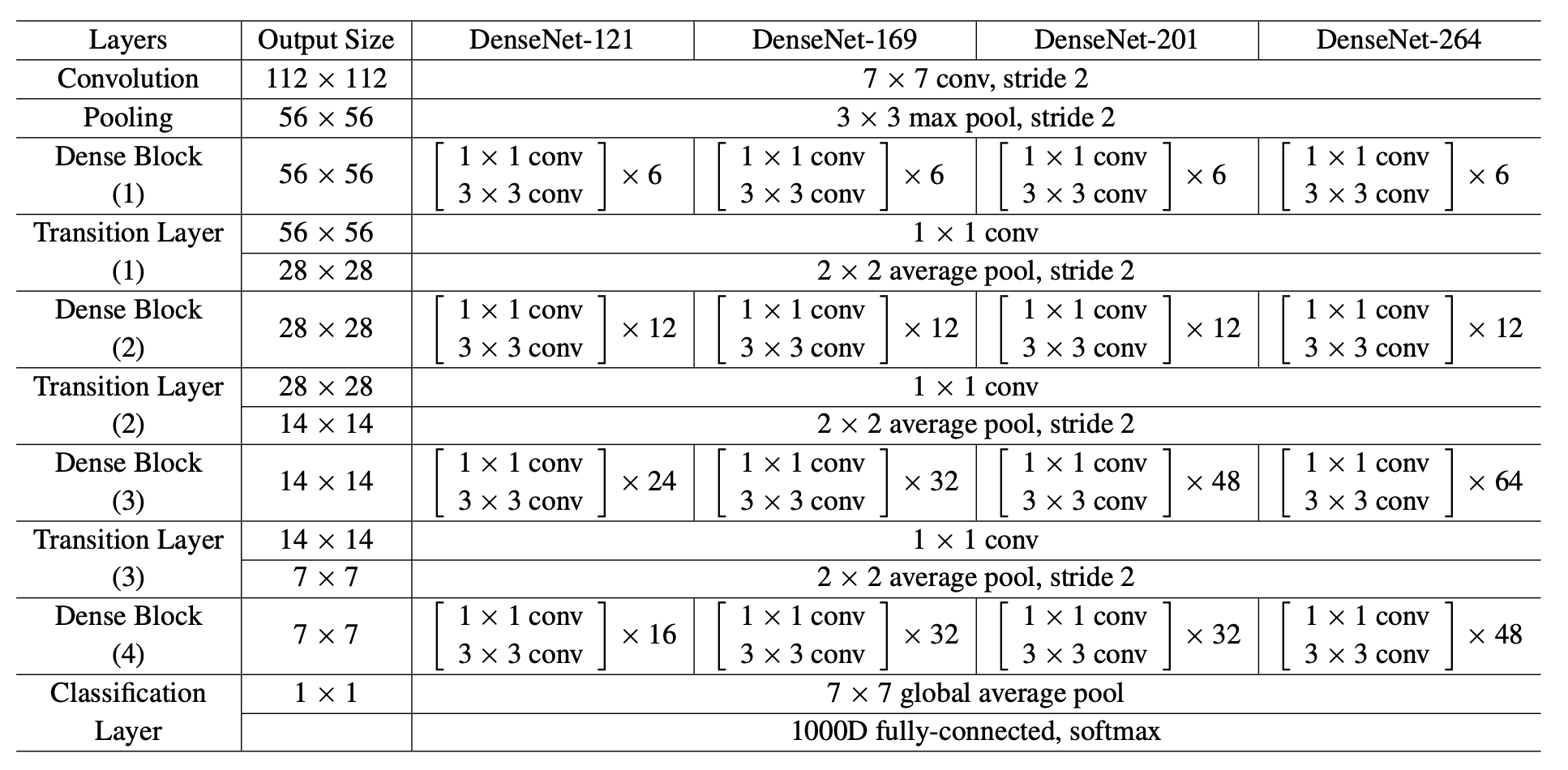}
     \caption{Dense Convolutional Network (DenseNet), connects each layer to every other layer in a feed-forward fashion\cite{amaarora_2020}.}
     \label{fig:firstExample}
\end{figure*}

\section{Results}
In this section, we would like to present our results with 5 test images (As shown in Fig.6.). On the leftmost column is our output image, and in the middle and right are the label image and the ground truth image respectively. The output image greatly satisfies the depth prediction shown by the label image, which shows that our model can achieve the depth prediction for indoor scenarios.

\section{Evaluation}

\subsection{Model size comparison.}
This section will compare the model size between our model with other models. As shown in Table 1, model sizes with MonoDepth and DenseNet are both greater than 160 megabytes, while our model is holding with the size of 144 megabytes. Smaller model sizes can expand the application area of the model, and we can deploy the lightweight model to platforms with fewer GPU resources (i.e. Mobile platforms). We will further discuss this part in the discussion section. Even with a smaller model size in our model, we still manage to maintain high accuracy. We will compare the accuracy of the model in detail in the next section. 

\subsection{Quantitative evaluation}
We will discuss the quantitative analysis of our model in this section. By comparing the quantitative results of five other different models as shown in Table 2, we compare their Root of the Mean of the Square of Errors(RMSE), Relative Squared Error (Sq Rel), and Mean of Absolute value of Errors (MAE). 

These error metrics are respectively defined as:

\[
    RMSE = \sqrt{\frac{1}{n} \sum^n_{i=1}(y_i - \hat{y_i})^2}
\]
\[
    SqRel = \frac{\sum^n_{i=1}(y_i - \hat{y_i})^2}{\sum^n_{i=1}(y_i - \bar{y_i})^2}
\]
where
\[
    \bar{y} = \frac{1}{n}\sum^n_{i=1}y_i
\]

\[
    MAE = \frac{1}{n}\sum_{i=1}^n |y_i - \hat{y_i}|
\]

RMSE quantifies the difference between the output and the label image. Compared to the other models, our model achieves a better result of 0.176 in RMSE (the smaller the value, the closer it is to the original image). Relative Squared Error scores 0.994 for the model. Finally, we obtain a smaller value for MAE than the other models, indicating that the average absolute distance between the pixels in the output image and their counterparts in the label image is smaller. These values prove that our model achieves good prediction results.

\begin{table}[]
    \centering
\caption{Model Size Comparison}
\label{quality}
\begin{tabular}{c|c|*{8}{c|}}
    \toprule
Model Name & Size \\
    \midrule
MonoDepth \cite{clement2018digging} & 343 MB\\
DenseNet \cite{Alhashim2018} & 165 MB\\
Ours Model  & 144 MB\\

\bottomrule
\end{tabular}
\vspace{-0.3cm}
\end{table}
\subsection{Limitation}
Due to the training machine's VRAM capacity, we can only train our model with less batch size. In this case, we may be unable to update the weight optimally. 
Since the dataset\cite{alhashim_2022_ialhashimdensedepth} we applied is based on a reduced version of the original NYU-V2 dataset, the smaller amount of data on the training scenario also affects the performance of the model. Meanwhile, the NYU dataset is mainly for indoor scenes, and for the outdoor scenario, we need to continue training our model based on the outdoor scenario dataset. We are also facing limitations from the network. As in the result shown in Fig.7. below, the model can generate inaccurate predictions from the image (i.e, the monitor section of the image)

\begin{figure}[htb]
     \centerline{
     \includegraphics[width=9cm]{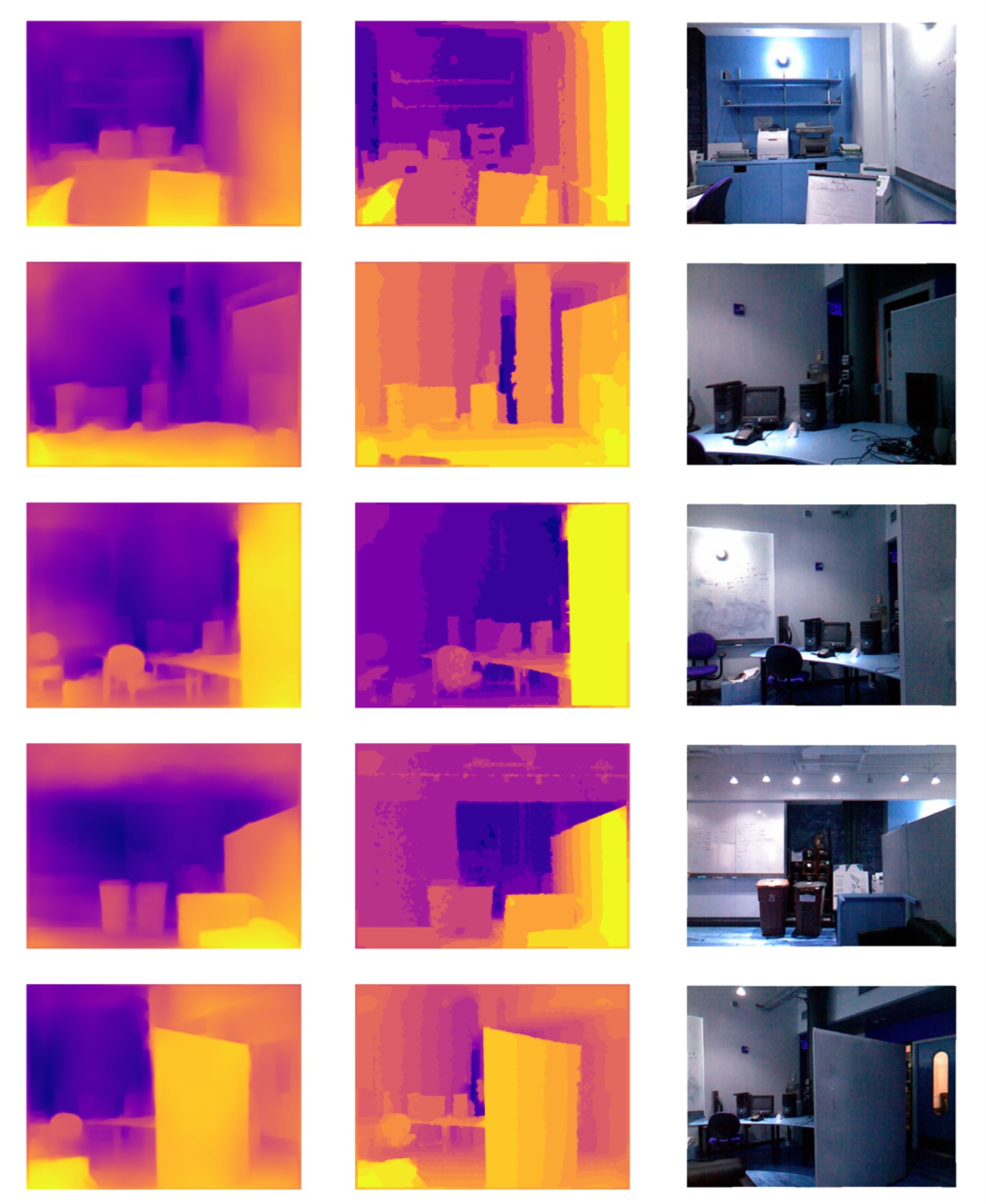}}
     \caption{Output images of the test result. The left most images are our result, with corresponding label images present in the middle. Right most is the input image}
     \label{testResult_1}
\end{figure}

\begin{table}[]
    \centering
\caption{Comparison of accuracy with other five similar models}
\label{quality}
\begin{tabular}{c|c|*{8}{c|}}
    \toprule
&RMSE & Sq Rel & MAE \\
    \midrule
Eigen et al.\cite{eigen2014} & 7.156 & 1.515 & -\\
Godard et al. \cite{godar2016}  & 4.935 & 8.898 & -\\
Kuznietsov et al. \cite{Kuznietsov2017}  & 4.621 & 0.741 & -\\
GoogLeNetV1\_ROB \cite{szegedy2014} &0.53 & 0.14 & 0.42 \\
CSWS-E\_ROB \cite{li2017}  & 0.31 & 0.06 & 0.24\\
DORN\_ROB\cite{fu2018}  & 0.29 & 0.06 & 0.22\\
Ours  & 0.176 & 0.994 & 0.150\\
\bottomrule
\end{tabular}
\vspace{-0.3cm}
\end{table}

\section{Discussion}
In this paper, we present a comprehensive literature review of recent research on deep learning-based monocular depth estimation and also introduce our lightweight UNet model with the DenseNet-121 encoder. We achieve a smaller model with better results by optimizing the number of encoder parameters.

\begin{figure}[htb]
     \centerline{
     \includegraphics[width=9.5cm]{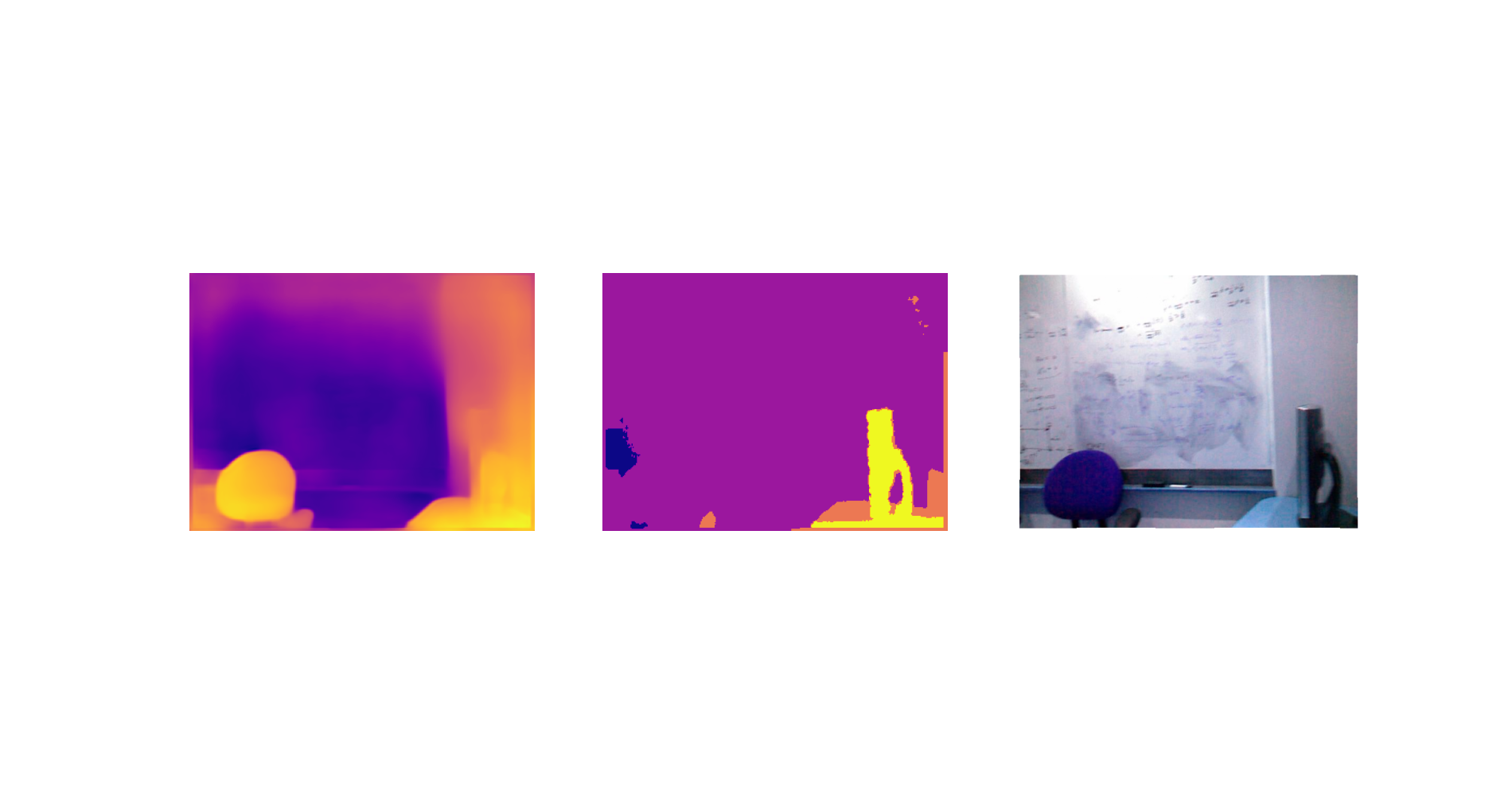}}
     \caption{An example of wrong depth estimation in a depth map. The left most images are our result, with corresponding label images present in the middle. Right most is the input image.}
     \label{testResult_1}
\end{figure}

In future works, we'll add weights to both $L1_{grad}$ and $L_{SSIM}$ loss function. We will also implement the transformer function in our network. We will develop a real-time depth estimate model, and integrate our model into mobile or smart home appliances like robot vacuums.
For mobile-end monocular image depth estimation, we are currently developing an Android application (Shown in the Fig.8.). We may import the model into the application and use the cloud GPU to produce the result using TensorFlow lite. Users can either upload local photographs that they have taken or take the photo by themselves into our UNet model for prediction. The corresponding result can be generated from the depth estimation image. 

\section{Conclusion}
As a conclusion of this paper, we learned about the existing monocular ranging methods through literature reviews and applied the deep learning structure of UNet in this method. We converted the monocular ranging problem to an image classification problem and applied DenseNet-121 to achieve the "goal of achieving the same depth prediction with fewer parameters." We demonstrate the effectiveness of our model by comparing and presenting the results of the output images. In addition, we summarize the impact of the existing hardware and structural limitations of UNet on the model, and we hope to expand the model further and apply it to more fields in future practice.

\begin{figure}[htb]
     \centerline{
     \includegraphics[width=10cm]{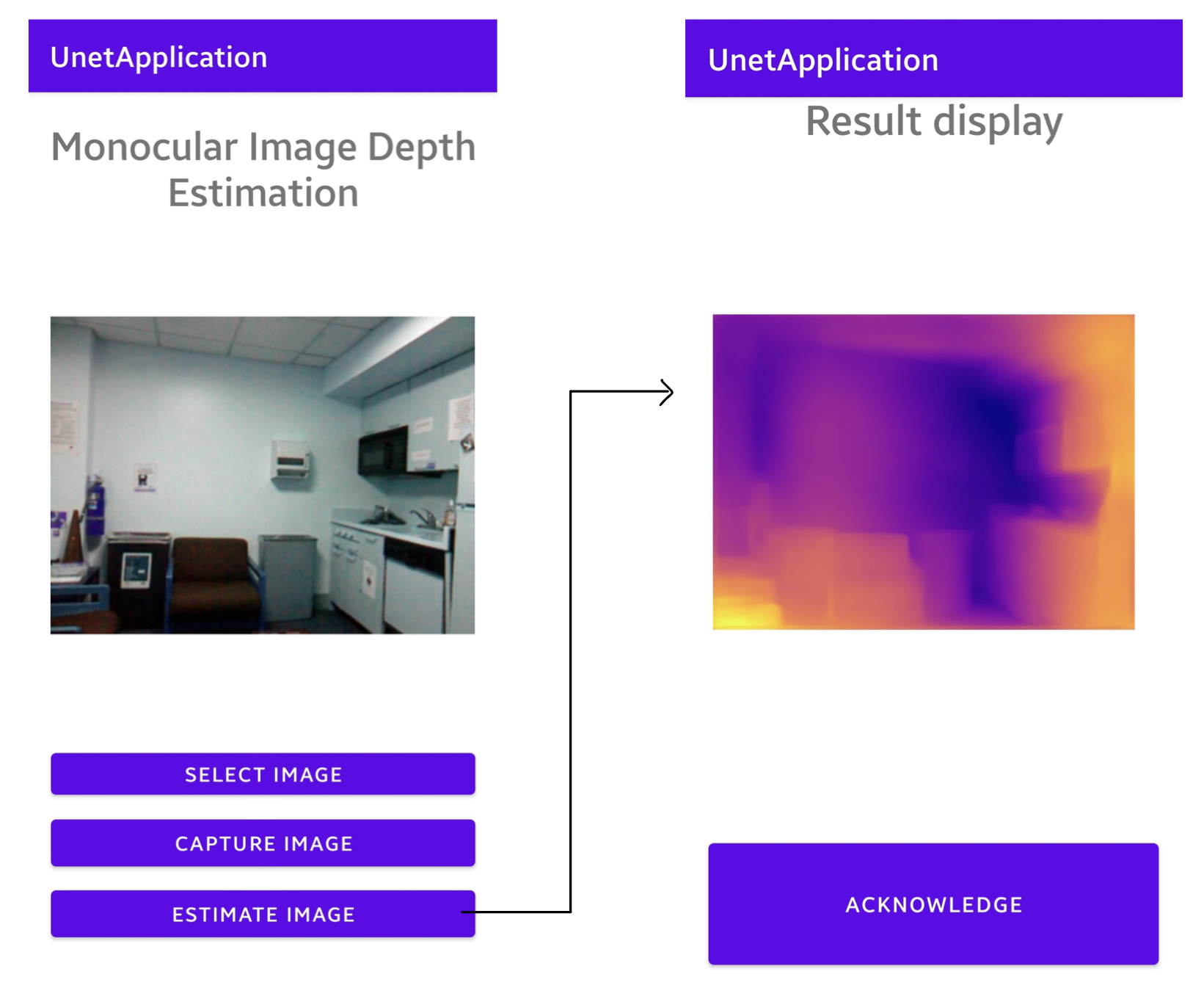}}
     \caption{An image shows the prototype of Android Application,  generating the depth estimation result from the user input image.}
     \label{testResult_1}
\end{figure}

\bibliographystyle{ieeetr}

\bibliography{egbibsample}

\end{document}